\documentclass{llncs}
\usepackage[utf8]{inputenc}
\usepackage{bm}
\usepackage{color}
\usepackage{subfig}
\usepackage{todonotes}
\usepackage{algorithm,graphicx}
\usepackage{multirow}
\usepackage{booktabs}
\usepackage{makecell}

\title{Unsupervised Domain Adaptation for Automatic Estimation of Cardiothoracic Ratio}
\author{Nanqing Dong
\inst{1,2} 
\and Michael Kampffmeyer
\inst{3}
\and Xiaodan Liang
\inst{4}
\and Zeya Wang
\inst{1}
\and
\\Wei Dai
\inst{1}
\and Eric Xing
\inst{1}
}
\institute{
Petuum, Inc., Pittsburgh, PA 15217, USA \\
\and
Cornell University, Ithaca, NY 14850, USA \\ 
\and
UiT The Arctic University of Norway, 9019 Tromsø, Norway \\
\and
Carnige Mellon University, Pittsburgh, PA 15213, USA \\
}

\begin{document}
\pagestyle{headings}
\maketitle
\begin{abstract}
The cardiothoracic ratio (CTR), a clinical metric of heart size in chest X-rays (CXRs), is a key indicator of cardiomegaly.
Manual measurement of CTR is time-consuming and can be affected by human subjectivity, making it desirable to design computer-aided systems that assist clinicians in the diagnosis process. Automatic CTR estimation through chest organ segmentation, however, requires large amounts of pixel-level annotated data, which is often unavailable.
To alleviate this problem, we propose an unsupervised domain adaptation framework based on adversarial networks. The framework learns domain invariant feature representations from openly available data sources to produce accurate chest organ segmentation for unlabeled datasets.
Specifically, we propose a model that enforces our intuition that prediction masks should be domain independent. Hence, we introduce a discriminator that distinguishes segmentation predictions from ground truth masks. 
We evaluate our system's prediction based on the assessment of radiologists and demonstrate the clinical practicability for the diagnosis of cardiomegaly.
We finally illustrate on the JSRT dataset that the semi-supervised performance of our model is also very promising.
\end{abstract}

\section{Introduction}
Cardiomegaly, also referred to as heart enlargement, is ranked as the most frequent disease code among a public collection of radiology reports from the National Library of Medicine (NLM) according to a National Institutes of Health (NIH) study on medical information retrieval \cite{demnerfushman2016preparing}. Cardiomegaly can result from other diseases or medical conditions, such as coronary artery disease and hypertension. It is suggested that cardiomegaly is associated with a high risk of sudden cardiac death \cite{tavora2012card}. The prevention of cardiomegaly starts from early detection and CTR measured from posterior-anterior (PA) CXR is an important indicator for cardiomegaly \cite{dimopoulos2011ctr}. 
CTR is calculated as the ratio of maximal horizontal cardiac diameter to maximal horizontal thoracic diameter, and CTR greater than 0.5 is commonly considered as cardiomegaly \cite{danzer1919ctr,dimopoulos2011ctr}. Manual measurement of CTR requires domain knowledge in radiology and extensive human labor in annotating CXRs, with results being error-prone due to observational error. This motivates the automation of CTR calculation and cardiomegaly detection. One common approach to estimating CTR is lung field segmentation~\cite{dallal2017automatic}. 

\begin{figure}[tbp]
\begin{center}
\includegraphics[width=0.8\linewidth]{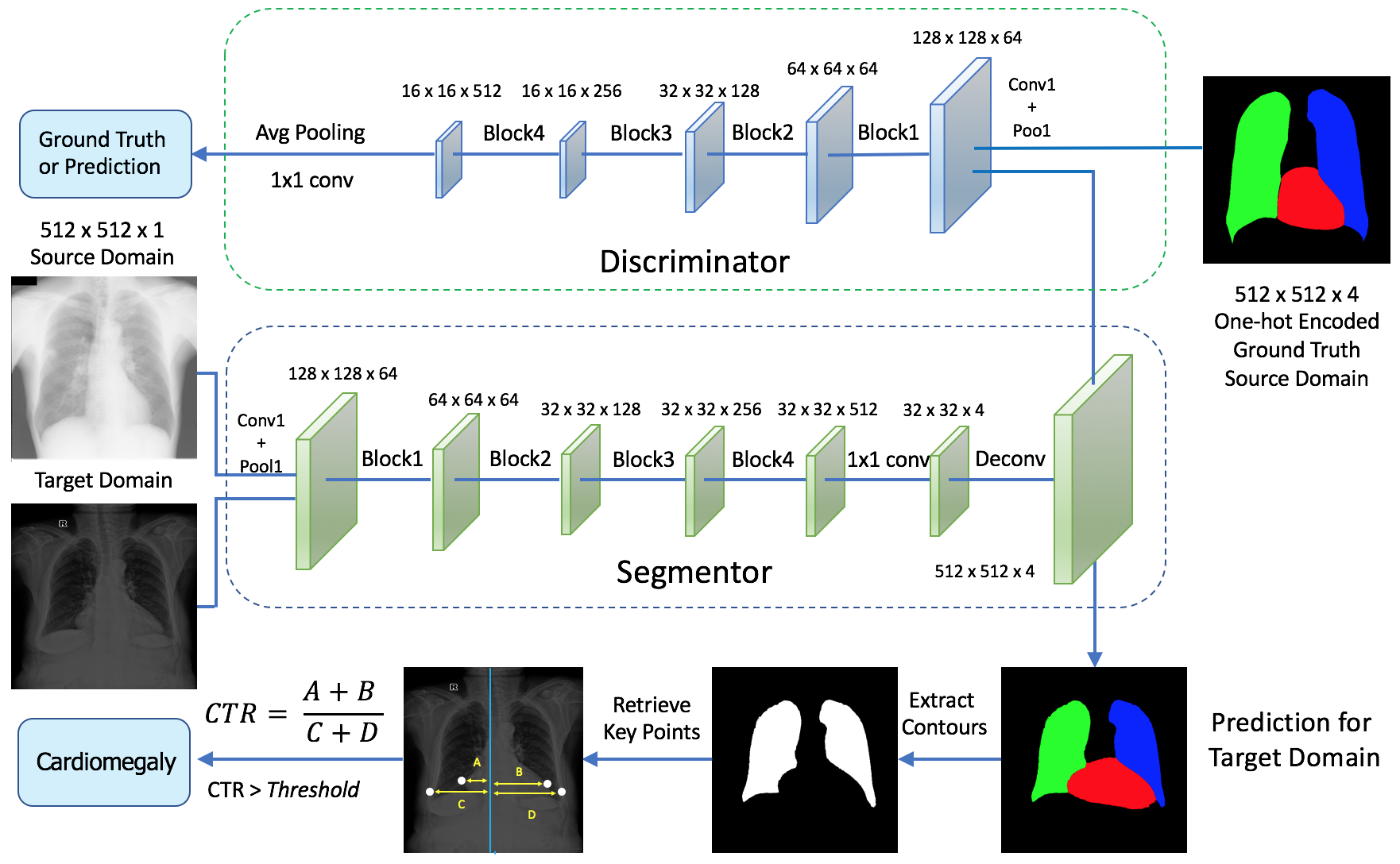}
\end{center}
   \caption{Illustration of the architecture. In our proposed adversarial training procedure, the segmentor produces segmentations for the input images and the discriminator attempts to distinguish these predictions from ground truth annotations. A post-processing step (bottom part of figure) is used to predict cardiomegaly based on the predicted lung segmentation masks.}
\label{fig:arch}
\end{figure}

Recent advances in Convolutional Neural Networks (CNNs) have brought breakthroughs in the field of semantic segmentation, achieving state-of-the-art performance \cite{chen2017deeplab,long2015fully}. Compared to traditional semantic segmentation, the annotated data for medical image segmentation is more difficult to be acquired, because of the limited available data and the tremendous cost of collecting and labeling it.
Transfer learning is a common approach to solve tasks with data scarcity, utilizing the fact that CNNs generally learn feature representations that are robust across a variety of tasks~\cite{tzeng2017adversarial}. However, as segmentation predictions based on these representations do not generalize very well to different datasets because of the dataset shift phenomena~\cite{5376}, it is commonly required to fine-tune the network based on a set of labels for the target domain. In particular, CXRs from different hospitals are often taken with different imaging protocols and commonly exhibit differences in noise levels, contrast and resolution. So it is impractical to directly use transfer learning techniques. See Figure~\ref{fig:arch} and Figure~\ref{fig:demo} for the differences between CXRs obtained at different hospitals.

In this paper, we propose an unsupervised domain adaptation (UDA) framework based on adversarial networks, which allows us to learn domain invariant feature representations from openly available data sources in order to produce accurate chest organ segmentation for unlabeled datasets.
Domain adaptation methods aim to reduce the problems of dataset shift, commonly, by aligning the learned source and target representation in a joint embedding space~\cite{shu2018a,tzeng2017adversarial}. Adversarial networks have become a popular choice to achieve this alignment, by introducing a discriminator that is trained to distinguish between the source and the target domain and by forcing the model to learn representations that can fool the discriminator. We propose an alternative training scheme where we utilize a discriminator that enforces our intuition that prediction masks should be domain independent by discriminating segmentation predictions from ground truth masks.
We evaluate our system's performance based on the assessment of radiologists on a CTR estimation dataset. Our approach outperforms the state-of-the-art UDA and shows the clinical practicability for the diagnosis of cardiomegaly.
We finally illustrate that our approach can also be used for semi-supervised chest organ segmentation of the JSRT benchmark dataset.

\section{Methodology}
The complete pipeline is shown in Figure~\ref{fig:arch}. The adversarial neural network consists of a discriminator and a segmentor. To demonstrate the generalization and simplicity of the methodology, we use ResNet18 as a backbone architecture \cite{he2016deep}. The discriminator is a standard ResNet classifier and the segmentor is inspired by the Fully Convolutional Network (FCN) \cite{long2015fully}, but uses an output stride of 16, following the example of \cite{chen2017deeplab}. 
Provided the predicted labels for the two lungs, the CTR is calculated in a post-processing step.

\subsection{Adversarial Training for Supervised Semantic Segmentation}
Adversarial learning was first introduced in the Generative Adversarial Network (GAN)~\cite{goodfellow2014generative} as a two-model zero-sum game, in which one model generates candidates for the other network to evaluate. Inspired by~\cite{luc2016semantic}, who used adversarial learning to improve semantic segmentation results, we let $S$ be the segmentor and $D$ be the discriminator. $S$ is trained to produce realistic prediction masks in order to fool $D$, which in turn is attempting to discriminate these predictions from ground truth images in a binary classification. $D$ is encouraged to learn a complex loss between the higher-order label statistics, which in practice cannot be explicitly formulated. Medical domain knowledge is being implicitly incorporated into this formulation as part of the annotated ground truth data.

An alternative training scheme is applied to train the segmentor and discriminator.
Given $D$, the loss to be minimized for $S$ is a multi-class cross-entropy loss for semantic segmentation, in addition to the binary cross-entropy loss for segmentation prediction $S(\bm{x})$ being classified as ground truth by $D$ \cite{luc2016semantic}.
\begin{equation}
    J_{seg}(S(\bm{x}),\bm{y}) = -\frac{1}{B_S}\sum_s \frac{1}{HW}\sum_{i}\sum_{c} y_{s,i,c}\log S(x_{s,i,c})
\label{eq:1}
\end{equation}
\begin{equation}
    J_{S}(S(\bm{x}),\bm{y}) = J_{seg}(S(\bm{x}),\bm{y}) - \lambda_{adv} \frac{1}{B_S} \sum_{s} \log D(S(x_s))
\label{eq:2}
\end{equation}
We use $x_s$ and $y_s$ to denote the input image and the ground truth, respectively, where $x_s$ is of shape $[H,W,1]$ and $y_s$ is of shape $[H,W,C]$ for $C$-class one-hot encoded labels. $B_S$ denotes the batch size for the segmentor training and $i$ ranges over all the spatial positions.
Given $S$, $D$ is optimized to maximize the probability of correctly distinguishing $S(\bm{x})$ from $\bm{y}$ as
\begin{equation}
J_{D}(S(\bm{x}),\bm{y}) = -\frac{1}{B_D}\sum_s \left[ \log(D(y_s)) + \log(1 - D(S(x_s)))\right] \; ,
\label{eq:3}
\end{equation}
where $B_D$ is the batch size for the discriminator training.

\subsection{Unsupervised Domain Adaption}
Our approach to unsupervised domain adaptation is illustrated in Figure~\ref{fig:arch} and is based on the idea that prediction masks, unlike input images and intermediate feature representations, can be considered domain independent. Unlike in~\cite{luc2016semantic}, we do not only make use of a discriminator to judge the quality of the segmentation mask, but also use it to align both source and target segmentation results with the domain-independent prediction mask.
We propose an alternative training scheme, where we present the discriminator with real ground truth images from our source domain, $y_s$, and with segmentation mask predictions from both the source and the target domain, $x_s$ and $x_t$, respectively. In order to learn domain invariant feature representations, we exploit the fact that we can train the segmentor using both the segmentation and the discriminator loss in the source domain to produce accurate segmentation prediction masks. However, simultaneously we enforce the fact that the segmentation masks for the target domain need to be of high quality. The updated losses are
\begin{equation}
    J_{S-DA}(S(\bm{x}),\bm{y}) = J_{S}(S(\bm{x}),\bm{y}) - \lambda_{adv} \frac{1}{B_S} \sum_{t} \log D(S(x_t)),
\label{eq:4}
\end{equation}

\begin{equation}
    J_{D-DA}(S(\bm{x}),\bm{y}) = J_{D}(S(\bm{x}),\bm{y}) - \frac{1}{B_D}\sum_t \log(1 - D(S(x_t))).
\label{eq:5}
\end{equation}

\subsection{Estimation of CTR}
\begin{figure}[tbp]
\begin{minipage}{0.49\columnwidth}
\includegraphics[width=\linewidth, height=\linewidth]{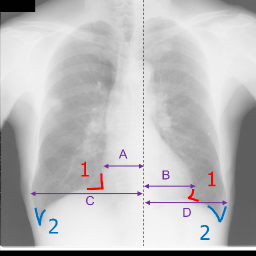}
\caption{Contour landmarks for lower lungs: cardiodiaphragmatic angles (1) and costophrenic angles (2).}
\label{fig:ctr}
\end{minipage}
\hfill
\begin{minipage}{0.49\columnwidth}
\captionsetup[subfigure]{labelformat=empty}
\subfloat{\includegraphics[width=0.32\linewidth]{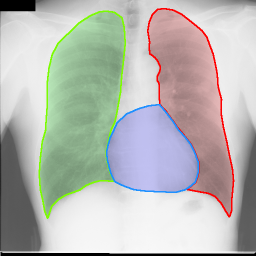}}\vspace{-0.05cm}
\subfloat{\includegraphics[width=0.32\linewidth]{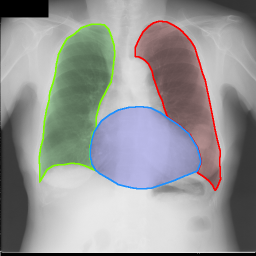}}\vspace{-0.05cm}
\subfloat{\includegraphics[width=0.32\linewidth]{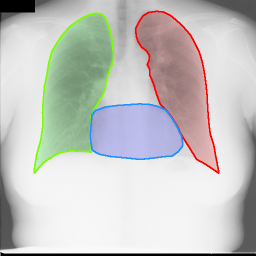}}\vspace{-0.05cm}\\
\subfloat{\includegraphics[width=0.32\linewidth,height=0.32\linewidth]{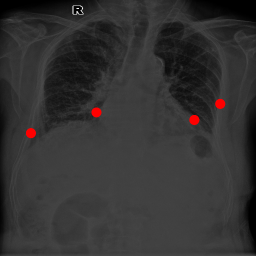}}\vspace{-0.05cm}
\subfloat{\includegraphics[width=0.32\linewidth,height=0.32\linewidth]{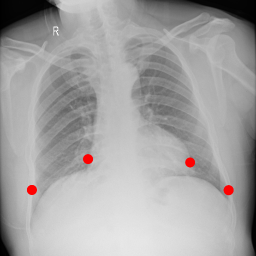}}\vspace{-0.05cm}
\subfloat{\includegraphics[width=0.32\linewidth,height=0.32\linewidth]{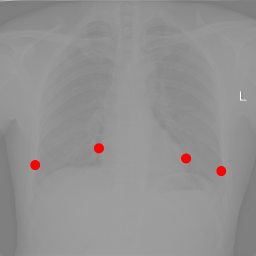}}\vspace{-0.05cm}
\caption{Example images of the two datasets. The three images in the top row correspond to examples of the JSRT dataset, overlaid with the segmentation annotation. The three images in the second row originate from the Wingspan dataset overlaid with the key points for the CTR calculation.}
\label{fig:demo}
\end{minipage}
\end{figure}

CTR is the ratio of maximal horizontal cardiac diameter to maximal horizontal thoracic diameter, as formulated in the Danzer Method~\cite{danzer1919ctr}. The diameters are the horizontal distance between horizontal coordinates of corresponding key points on the lung contours. As shown in Figure~\ref{fig:ctr}, the maximal horizontal cardiac diameter and maximal horizontal thoracic diameter can only be achieved by points above cardiodiaphragmatic angles and costophrenic angles, which can be retrieved by the use of a convex hull algorithm. With a hypothetical central line, the Danzer Method could be reinterpreted as $\frac{A+B}{C+D}$, while line segments A, B, C, D are all maximized independently. The constraints of maximizing $A+B$ are that the points of intersection between lung contours and A and B must be above cardiodiaphragmatic angles. The points of intersection between the lung contours and the maximized A, B, C, and D are the key points. Provided the estimated CTR, cardiomegaly can be predicted under different thresholds for different age groups. Following~\cite{dallal2017automatic}, the threshold, $T$, is chosen to be 0.5.

\subsection{Semi-Supervised Semantic Segmentation}
We further illustrate our model's ability for the task of semi-supervised learning. As the annotated data are limited, it is common in medical image segmentation to have only a subset of training data labeled. Provided with a set of labeled and unlabeled datapoints \{\{($x_1$, $y_1$),...,($x_l$,$y_l$)\},\{$\tilde{x}_1$,...,$\tilde{x}_u$\}\}, the task of semi-supervised learning aims to exploit the underlying data properties of the unlabeled data in addition to the labeled data. $l$ and $u$ correspond to the number of labeled and unlabeled examples, respectively.
Similar to our unsupervised domain adaptation, we adopt an alternating training strategy, where the model is presented with both labeled and unlabeled data. 
We optimize $S$ and $D$ using Equation~\ref{eq:4} and~\ref{eq:5} and treat the labeled data as the source domain and the unlabeled data as the target domain.
This lets us leverage the unlabeled data to align the distribution of segmentation predictions with the distribution of ground truth labels, effectively regularizing the model and improving overall performance.

\section{Experimental Results}
The {\bf JSRT} dataset is released by the Japanese Society of Radiological Technology (JSRT) \cite{jsrt} and is a benchmark dataset for lung and heart segmentation. JSRT contains 247 grayscale CXRs with annotated lung and heart pixel-wise labels, where 154 have lung nodules and 93 don't have lung nodules. Each CXR has a size of $2048 \times 2048$ and the pixel spacing is 0.175mm. In this paper, JSRT is used as the source domain for the unsupervised domain adaption. See Figure~\ref{fig:demo} for examples from the dataset overlaid with the ground truth annotation.

\begin{figure}[t]
\begin{center}
\captionsetup[subfigure]{labelformat=empty}
\subfloat{\includegraphics[width=0.14\linewidth,height=0.14\linewidth]{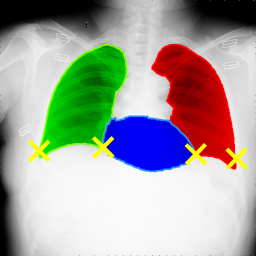}}\vspace{-0.05cm}
\subfloat{\includegraphics[width=0.14\linewidth,height=0.14\linewidth]{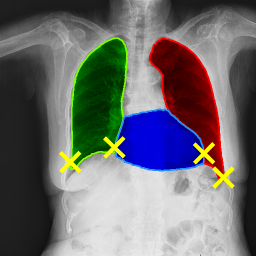}}\vspace{-0.05cm}
\subfloat{\includegraphics[width=0.14\linewidth,height=0.14\linewidth]{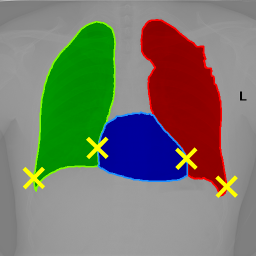}}\vspace{-0.05cm}
\subfloat{\includegraphics[width=0.14\linewidth,height=0.14\linewidth]{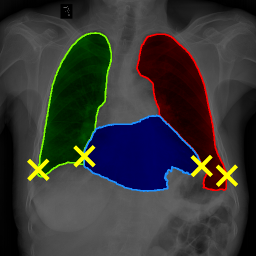}}\vspace{-0.05cm}
\subfloat{\includegraphics[width=0.14\linewidth,height=0.14\linewidth]{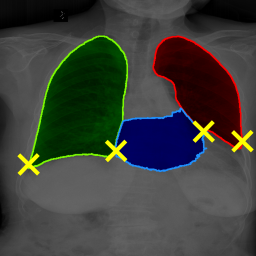}}\vspace{-0.05cm}
\subfloat{\includegraphics[width=0.14\linewidth,height=0.14\linewidth]{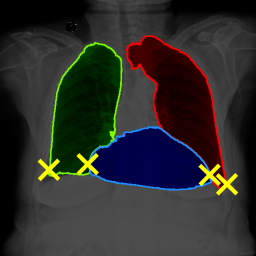}}\vspace{-0.05cm}
\end{center}
   \caption{Visualization of the segmentation and key point results for the Wingspan dataset for our proposed domain adaptation method.}
\label{fig:dares}
\end{figure}

The {\bf Wingspan} dataset is provided by a private research institute, Wingspan Technology. The dataset contains 221 grayscale CXRs for adult patients with annotated key points for calculation of CTR. Each image was annotated by two licensed radiologists independently, and the annotations were accepted by both annotators and an independent reviewer. The de-identified data were collected from 6 hospitals, which have different imaging protocols. The image sizes, pixel spacing and clinical setup vary for each CXR. See Figure~\ref{fig:demo} for examples from the dataset with key point annotations and the differences to the JSRT dataset and Figure~\ref{fig:dares} for the large variety in the data modalities, which is not present in the available public benchmark datasets.

In our work, we use the Wingspan dataset as the target domain. 
We investigate the potential of our proposed approach for unsupervised domain adaptation for the task of CTR estimation. For this, we utilize the segmentation masks of the source domain (JSRT) to perform segmentation on our target domain (Wingspan) and use the predicted segmentation result to compute the CTR.
We then show how our method can be easily adapted to semi-supervised semantic segmentation. We evaluate our approach on JSRT and illustrate that we can use the information encoded in our unlabeled data. 
The adversarial networks are trained using the Adam optimizer with a learning rate of $10^{-3}$. The discriminator is updated twice before the segmentor is updated, and $\lambda_{adv}$ is $10^{-4}$. We use $B_S = B_D = 8$. JSRT is randomly split into $80\%$ for training and $20\%$ for testing. For all the experiments in this paper, no data augmentation is used, which further shows the robustness of our approach. 

\begin{table}[tbp]
\centering
{\setlength{\tabcolsep}{0.4em}
\begin{tabular}{lccc}
\hline
Method & APE & MAE & RMSE \\ \Xhline{4\arrayrulewidth}%\hline
TL-SEG & $16.0\%\pm16.1\%$ & $8.9\%\pm9.3\%$ & $0.13$ \\ \hline
TL-ADV & $11.4\%\pm11.2\%$ & $5.9\%\pm5.9\%$ & $0.08$ \\ \hline
ADDA & $9.2\%\pm9.9\%$ & $5.1\%\pm5.8$ & $0.08$ \\ \hline
DA-ADV & $5.8\%\pm8.5\%$ & $3.3\%\pm5.1\%$ & $0.06$ \\ \hline
\end{tabular}
}
\caption{Results for the unsupervised domain adaptation of CTR estimation experiments. APE denotes average percentage error, MAE denotes mean absolute error, and RMSE denotes root mean square error.}
\label{tab:da}
\end{table}

\paragraph{\bf{Unsupervised Domain Adaptation:}}
To assess our performance for unsupervised domain adaptation, we compare our approach (DA-ADV) to three alternative approaches and present the quantitative results for the CTR estimation in Table~\ref{tab:da}.
The baseline uses the segmentor trained on the source domain directly on the target domain. This corresponds to transfer learning without fine-tuning on the target domain (TL-SEG). The baseline segmentor can be improved by adding a discriminator with an adversarial training scheme (TL-ADV). Finally, we compare with one of the state-of-the-art approaches for domain adaptation, ADDA \cite{tzeng2017adversarial}, which trains a segmentation network and then utilizes an adversarial loss to align the source and the target domain feature representations in order to minimize data shift. However, ADDA's performance is highly dependent on the quality of the segmentation network, which is not robust. We observe that our method outperforms the alternative approaches, providing considerable improvements for CTR estimation.
Qualitative results for the predicted segmentation masks and the key points for images from the Wingspan dataset can be seen in~Figure~\ref{fig:dares}.
Based on the threshold of 0.5, we predict cardiomegaly with our pipeline and achieve $87.78\%$ in accuracy, $97.72\%$ in precision, $84.21\%$ in sensitivity and $95.57\%$ in specificity.

\begin{table}[t]
\centering
{\setlength{\tabcolsep}{0.4em}
\begin{tabular}{lccc}
\hline
Method & IoU (Lungs) & IoU (Heart) \\ \Xhline{4\arrayrulewidth}%\hline
Human Observer~\cite{van2006segmentation} & $94.6\%\pm1.8\%$ & $87.8\%\pm5.4\%$ \\ \hline
Supervised & $95.5\%\pm 0.3\%$ & $90.2\%\pm 0.5\%$ \\ \hline
Supervised ($50\%$) & $82.9\%\pm 3.5\%$ & $71.2\%\pm 7.6\%$ \\ \hline
Supervised ($25\%$) & $75.4\%\pm 5.7\%$ & $62.4\%\pm 11.9\%$ \\ \hline
Supervised ($10\%$) & $60.1\%\pm 9.6\%$ & $39.4\%\pm 14.7\%$ \\ \hline
Semi-Supervised ($50\%$) & $90.4\%\pm 3.1\%$ & $81.2\%\pm 2.5\%$ \\ \hline
Semi-Supervised ($25\%$) & $89.9\%\pm 3.3\%$ & $75.5\%\pm 5.4\%$ \\ \hline
Semi-Supervised ($10\%$) & $81.7\%\pm 4.6\%$ & $69.4\%\pm 7.2\%$ \\ \hline
\end{tabular}
}
\caption{Results for the semi-supervised segmentation experiments. IoU denotes the Intersection over Union.}
\label{tab:semi}
\end{table}
\paragraph{\bf{Semi-Supervised Semantic Segmentation:}}
As a baseline we train the model respectively on $10\%$, $25\%$ and $50\%$ of annotated data in a supervised manner. As a comparison, we train the model on the whole dataset in a semi-supervised manner, while only portions of the data used in the supervised setting are provided with the labels. 
Table~\ref{tab:semi} provides the results of our semi-supervised experiments.
Our approach clearly makes use of the unlabeled data, achieving large performance gains. To put our results into perspective and to illustrate the performance that can be achieved when all training labels are available, we also train the model on the fully labeled training dataset.

\section{Conclusions}
In this paper, we present an approach to unsupervised domain adaptation for the task of CTR estimation that is based on the intuition that prediction masks should be domain independent. Using an adversarial training approach, we show that we can predict cardiomegaly from a dataset without segmentation annotations. We further illustrate how our approach can be adapted for semi-supervised learning.

\vspace{0.3cm}
\noindent{\bf{Acknowledgements.}} We thank Wingspan Technology for collecting and annotating the data for this study.

\bibliographystyle{splncs03}
\bibliography{miccai}

\begin{thebibliography}{10}
\providecommand{\url}[1]{\texttt{#1}}
\providecommand{\urlprefix}{URL }

\bibitem{chen2017deeplab}
Chen, L., Papandreou, G., Kokkinos, I., Murphy, K., Yuille, A.: Deeplab:
  Semantic image segmentation with deep convolutional nets, atrous convolution,
  and fully connected crfs. IEEE Transactions on Pattern Analysis and Machine
  Intelligence  (2017)

\bibitem{dallal2017automatic}
Dallal, A.H., Agarwal, C., Arbabshirani, M.R., Patel, A., Moore, G.: Automatic
  estimation of heart boundaries and cardiothoracic ratio from chest x-ray
  images. In: Medical Imaging 2017: Computer-Aided Diagnosis. vol. 10134, p.
  101340K. International Society for Optics and Photonics (2017)

\bibitem{danzer1919ctr}
Danzer, C.S.: The cardiothoracic ratio. The American Journal of the Medical
  Sciences  157,  513--554 (1919)

\bibitem{demnerfushman2016preparing}
Demner-Fushman, D., Kohli, M.D., Rosenman, M.B., Shooshan, S.E., Rodriguez, L.,
  Antani, S.K., Thoma, G.R., McDonald, C.J.: Preparing a collection of
  radiology examinations for distribution and retrieval. Journal of the
  American Medical Informatics Association : JAMIA  23 2,  304--10 (2016)

\bibitem{dimopoulos2011ctr}
Dimopoulos, K., Giannakoulas, G., Bendayan, I., Liodakis, E., Petraco, R.,
  Diller, G.P., Piepoli, M., Swan, L., Mullen, M., Best, N., A~Poole-Wilson,
  P., Francis, D., Rubens, M., A~Gatzoulis, M.: Cardiothoracic ratio from
  postero-anterior chest radiographs: A simple, reproducible and independent
  marker of disease severity and outcome in adults with congenital heart
  disease. International Journal of Cardiology  166 (2011)

\bibitem{goodfellow2014generative}
Goodfellow, I., Pouget-Abadie, J., Mirza, M., Xu, B., Warde-Farley, D., Ozair,
  S., Courville, A., Bengio, Y.: Generative adversarial nets. In: Advances in
  Neural Information Processing Systems. pp. 2672--2680 (2014)

\bibitem{5376}
Gretton, A., Smola, A., Huang, J., Schmittfull, M., Borgwardt, K.,
  Sch{\"o}lkopf, B.: Covariate shift and local learning by distribution
  matching, pp. 131--160. MIT Press, Cambridge, MA, USA (2009)

\bibitem{he2016deep}
He, K., Zhang, X., Ren, S., Sun, J.: Deep residual learning for image
  recognition. In: Proceedings of the IEEE Conference on Computer Vision and
  Pattern Recognition. pp. 770--778 (2016)

\bibitem{long2015fully}
Long, J., Shelhamer, E., Darrell, T.: Fully convolutional networks for semantic
  segmentation. In: Proceedings of the IEEE Conference on Computer Vision and
  Pattern Recognition. pp. 3431--3440 (2015)

\bibitem{luc2016semantic}
Luc, P., Couprie, C., Chintala, S., Verbeek, J.: Semantic segmentation using
  adversarial networks. In: Advances in Neural Information Processing Systems
  \textit{Adversarial Training Workshop} (2016)

\bibitem{jsrt}
Shiraishi, J., Katsuragawa, S., Ikezoe, J., Matsumoto, T., Kobayashi, T.,
  Komatsu, K.i., Matsui, M., Fujita, H., Kodera, Y., Doi, K.: Development of a
  digital image database for chest radiographs with and without a lung nodule:
  receiver operating characteristic analysis of radiologists' detection of
  pulmonary nodules. American Journal of Roentgenology  174(1),  71--74 (2000)

\bibitem{shu2018a}
Shu, R., Bui, H., Narui, H., Ermon, S.: A {DIRT}-t approach to unsupervised
  domain adaptation. In: Internatioanl Conference on Learning Representation
  (2018)

\bibitem{tavora2012card}
Tavora, F., Zhang, Y., Zhang, M., Li, L., Ripple, M., Fowler, D., Burke, A.:
  Cardiomegaly is a common arrhythmogenic substrate in adult sudden cardiac
  deaths, and is associated with obesity. Pathology  44,  187--91 (03 2012)

\bibitem{tzeng2017adversarial}
Tzeng, E., Hoffman, J., Saenko, K., Darrell, T.: Adversarial discriminative
  domain adaptation. In: Proceedings of the IEEE Conference on Computer Vision
  and Pattern Recognition. pp. 2962--2971 (2017)

\bibitem{van2006segmentation}
Van~Ginneken, B., Stegmann, M.B., Loog, M.: Segmentation of anatomical
  structures in chest radiographs using supervised methods: a comparative study
  on a public database. Medical Image Analysis  10(1),  19--40 (2006)

\end{thebibliography}

\title{Unsupervised Domain Adaptation for Automatic Estimation of Cardiothoracic Ratio Supplementary Material}
\author{}
\institute{}
\maketitle

\section{Additional visualizations}
\begin{figure}[H]
\centering{
\captionsetup[subfigure]{labelformat=empty}
\subfloat{\includegraphics[width=0.23\linewidth]{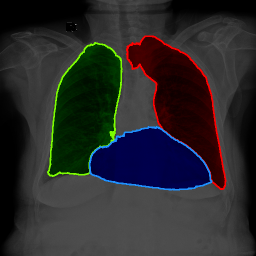}}\vspace{-0.05cm}
\subfloat{\includegraphics[width=0.23\linewidth]{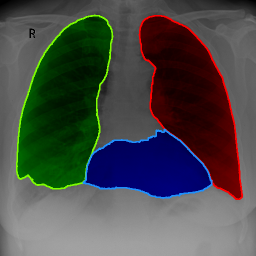}}\vspace{-0.05cm}
\subfloat{\includegraphics[width=0.23\linewidth]{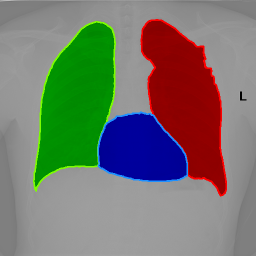}}\vspace{-0.05cm}
\subfloat{\includegraphics[width=0.23\linewidth]{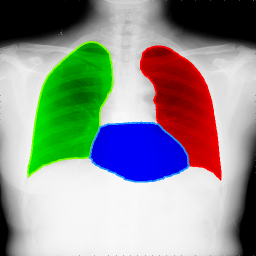}}\vspace{-0.05cm}\\
\subfloat{\includegraphics[width=0.23\linewidth]{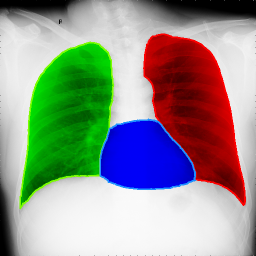}}\vspace{-0.05cm}
\subfloat{\includegraphics[width=0.23\linewidth]{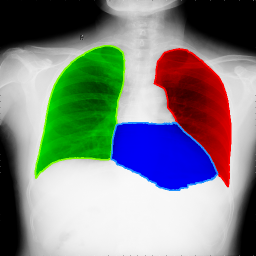}}\vspace{-0.05cm}
\subfloat{\includegraphics[width=0.23\linewidth]{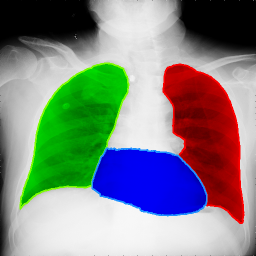}}\vspace{-0.05cm}
\subfloat{\includegraphics[width=0.23\linewidth]{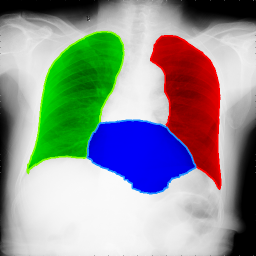}}\vspace{-0.05cm}\\
\subfloat{\includegraphics[width=0.23\linewidth]{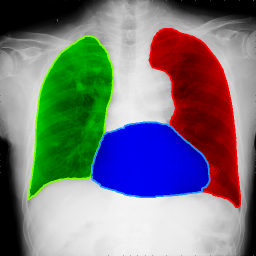}}\vspace{-0.05cm}
\subfloat{\includegraphics[width=0.23\linewidth]{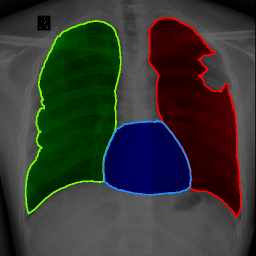}}\vspace{-0.05cm}
\subfloat{\includegraphics[width=0.23\linewidth]{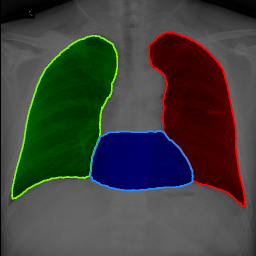}}\vspace{-0.05cm}
\subfloat{\includegraphics[width=0.23\linewidth]{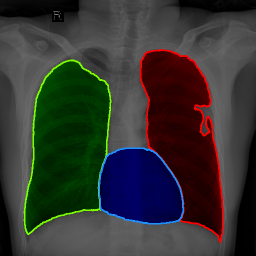}}\vspace{-0.05cm}\\
\subfloat{\includegraphics[width=0.23\linewidth]{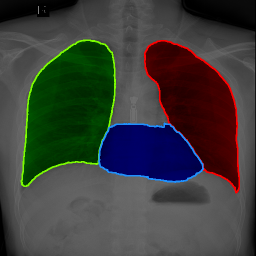}}\vspace{-0.05cm}
\subfloat{\includegraphics[width=0.23\linewidth]{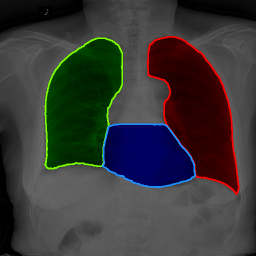}}\vspace{-0.05cm}
\subfloat{\includegraphics[width=0.23\linewidth]{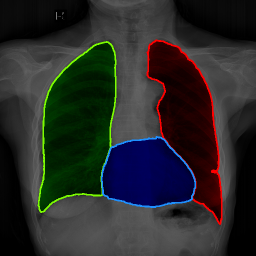}}\vspace{-0.05cm}
\subfloat{\includegraphics[width=0.23\linewidth]{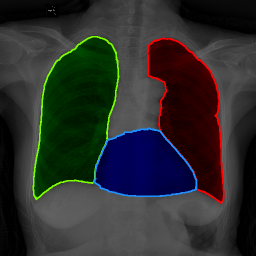}}\vspace{-0.05cm}
}
\caption{Segmentation results for our domain adaptation method on the Wingspan target domain dataset.}
\label{fig:res}
\end{figure}

In Figure~\ref{fig:res} we provide some more and enlarged qualitative examples for the segmentation results obtained on the Wingspan dataset as part of our unsupervised domain adaptation approach.

\begin{figure}[H]
\centering{
\captionsetup[subfigure]{labelformat=empty}
\subfloat{\includegraphics[width=0.23\linewidth]{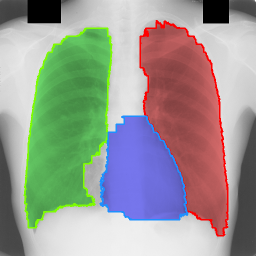}}\vspace{-0.05cm}
\subfloat{\includegraphics[width=0.23\linewidth]{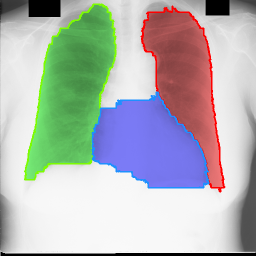}}\vspace{-0.05cm}
\subfloat{\includegraphics[width=0.23\linewidth]{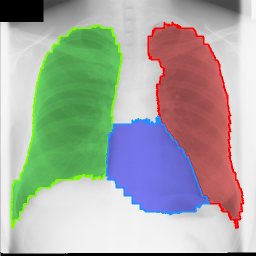}}\vspace{-0.05cm}
\subfloat{\includegraphics[width=0.23\linewidth]{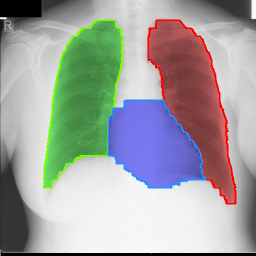}}\vspace{-0.05cm}\\
\subfloat{\includegraphics[width=0.23\linewidth]{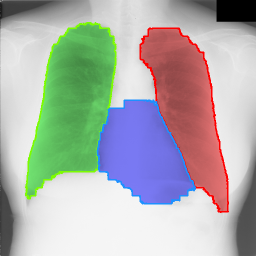}}\vspace{-0.05cm}
\subfloat{\includegraphics[width=0.23\linewidth]{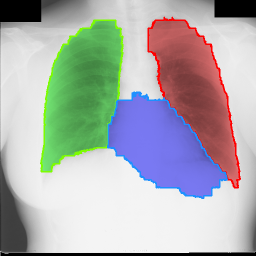}}\vspace{-0.05cm}
\subfloat{\includegraphics[width=0.23\linewidth]{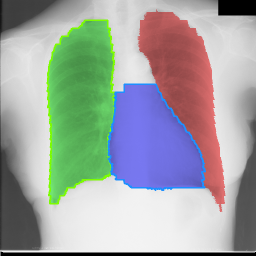}}\vspace{-0.05cm}
\subfloat{\includegraphics[width=0.23\linewidth]{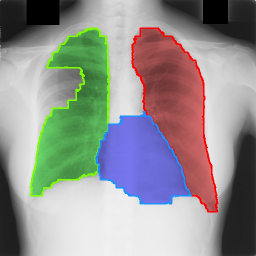}}\vspace{-0.05cm}\\
\subfloat{\includegraphics[width=0.23\linewidth]{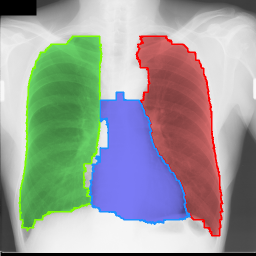}}\vspace{-0.05cm}
\subfloat{\includegraphics[width=0.23\linewidth]{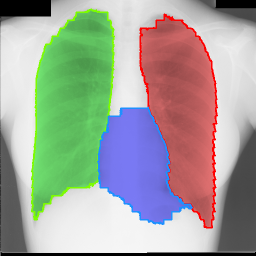}}\vspace{-0.05cm}
\subfloat{\includegraphics[width=0.23\linewidth]{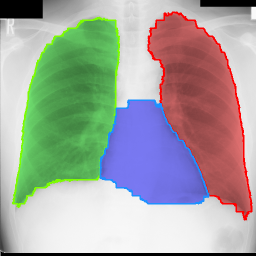}}\vspace{-0.05cm}
\subfloat{\includegraphics[width=0.23\linewidth]{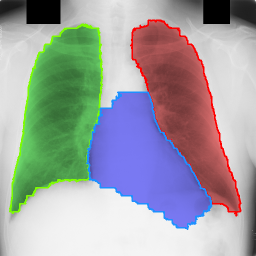}}\vspace{-0.05cm}\\
\subfloat{\includegraphics[width=0.23\linewidth]{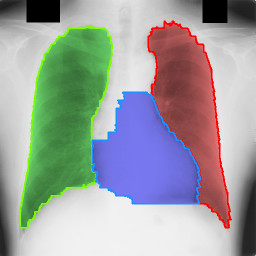}}\vspace{-0.05cm}
\subfloat{\includegraphics[width=0.23\linewidth]{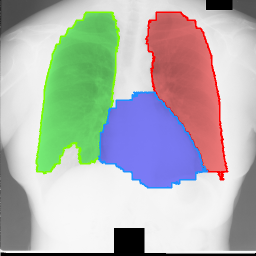}}\vspace{-0.05cm}
\subfloat{\includegraphics[width=0.23\linewidth]{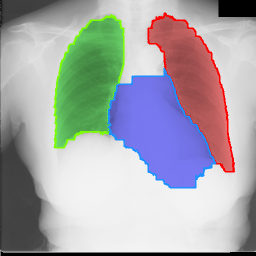}}\vspace{-0.05cm}
\subfloat{\includegraphics[width=0.23\linewidth]{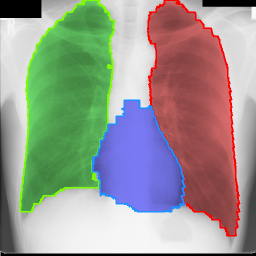}}\vspace{-0.05cm}
}
\caption{Segmentation results for the JSRT dataset when using semi-supervised training with $75\%$ of the labeled training data held-out.}
\label{fig:res2}
\end{figure}

In Figure~\ref{fig:res2} we provide some qualitative examples for the semi-supervised segmentation results obtained on the JSRT dataset.

\end{document}